\documentclass[11pt]{article}

\usepackage[final]{acl}

\usepackage{times}
\usepackage{latexsym}

\usepackage[T1]{fontenc}

\usepackage[utf8]{inputenc}

\usepackage{microtype}

\usepackage{inconsolata}

\usepackage{graphicx}
\usepackage{booktabs,makecell,xcolor}
\usepackage{multirow}
\usepackage{kotex}
\usepackage{url}
\usepackage{threeparttable}

\makeatletter
\def\@fnsymbol#1{\ensuremath{
  \ifcase#1\or *\or \dagger\or \ddagger\or \S\or \P\or \|\or **\or \dagger\dagger\or \ddagger\ddagger\fi}
}
\makeatother

\title{Korean Canonical Legal Benchmark: Toward Knowledge-Independent Evaluation of LLMs' Legal Reasoning Capabilities}

\author{
  Hongseok Oh{$^{*}$} \\
  University of Seoul \\
  \texttt{cxv0519@uos.ac.kr}
  \And
  Wonseok Hwang\thanks{Most work was done while at LBOX.}{$^{,\; \dagger}$} \\
  University of Seoul \\
  \texttt{wonseok.hwang@uos.ac.kr}
  \And
  Kyoung-Woon On\thanks{Corresponding authors.} \\
  LBOX \\
  \texttt{kyoungwoon.on@lbox.kr}
}

\begin{document}
\maketitle
\begin{abstract}

We introduce the \textbf{Korean Canonical Legal Benchmark (KCL)}, a benchmark designed to assess language models’ legal reasoning capabilities independently of domain-specific knowledge. KCL provides question-level supporting precedents, enabling a more faithful disentanglement of reasoning ability from parameterized knowledge. KCL consists of two components: (1) \textbf{KCL-MCQA}, multiple-choice problems of 283 questions with 1,103 aligned precedents, and (2) \textbf{KCL-Essay}, open-ended generation problems of 169 questions with 550 aligned precedents and 2,739 instance-level rubrics for automated evaluation. Our systematic evaluation of 30+ models shows large remaining gaps, particularly in KCL-Essay, and that reasoning-specialized models consistently outperform their general-purpose counterparts.
We release all resources, including the benchmark dataset and evaluation code, at \url{https://github.com/lbox-kr/kcl}.

\end{abstract}

\section{Introduction}

\begin{figure*}[tb!]
  \centering
  \includegraphics[width=\linewidth]{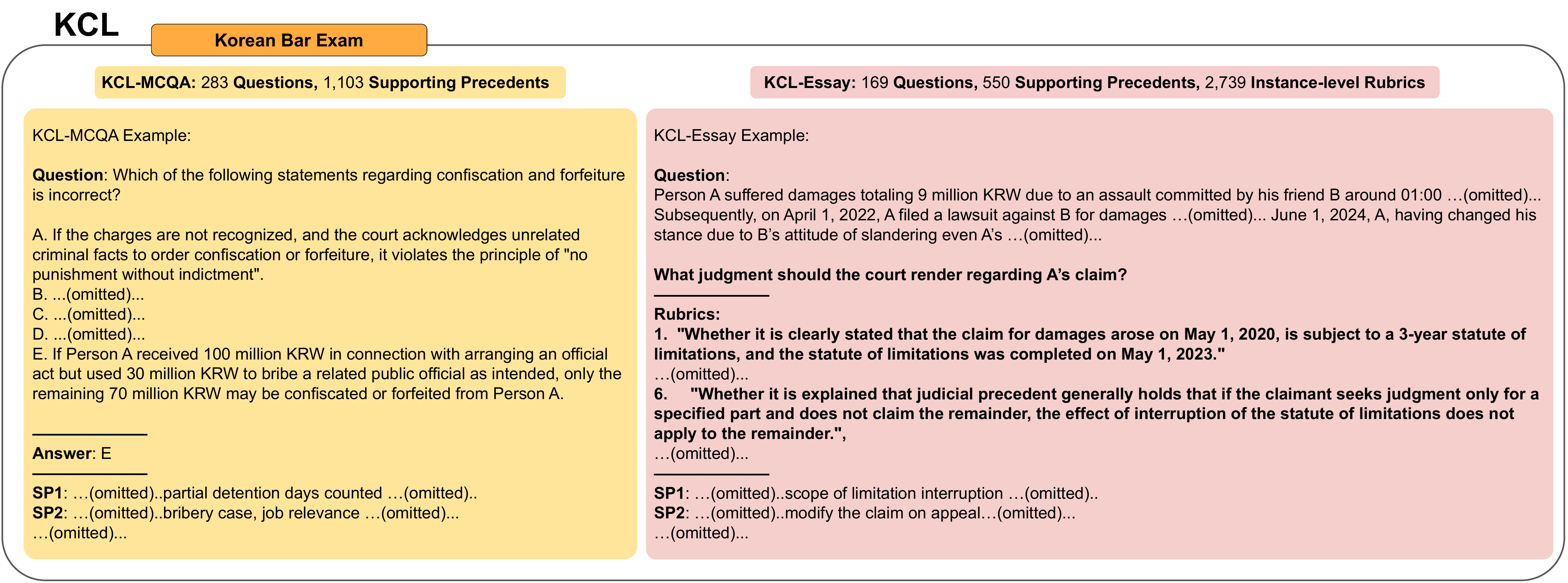}
  \caption{An overview of the KCL benchmark and representative samples translated into English.}
  \label{fig:KCL}
\vspace{-10pt}
\end{figure*}

Large reasoning models (LRMs) trained to reason explicitly in the verbal space have shown superior performance over general large language models~\cite{r1, o3}.
However, their capabilities have been evaluated mainly in formally structured domains such as mathematics, STEM, and code~\cite{gsm8k, gpqa, livecodebench}. 
A recent knowledge-independent, rule-following benchmark takes a step toward assessing LRMs’ reasoning independently of background knowledge~\cite{korbench}.

However, real-world applications require solving problems in vertical domains such as biomedicine, finance, and law, where the types of reasoning are diverse and context dependent.
In these domains, it is more challenging to isolate reasoning ability from the memorized knowledge.
Tasks in the legal domain are representative cases in which domain knowledge and logical reasoning are tightly intertwined, requiring complex context understanding and reasoning~\cite{verifiable}.
However, the need for country-specific expertise makes legal data less accessible for LLM development.
Consequently, prior benchmarks mainly focused on knowledge itself~\citep{lexglue,lboxopen,lawbench}.

Meanwhile, several recent benchmarks attempt to incorporate legal reasoning more explicitly, but remain limited along a few common dimensions.
\citet{legalbench} define tasks aligned with the Issue, Rule, Application, Conclusion (IRAC) framework and evaluate them using exact match, but are restricted to English-language contexts and do not provide supporting knowledge.
Similarly, \citet{lexam} construct IRAC-based tasks from English and German law exam questions and adopt a reference-based LLM-as-a-Judge evaluation scheme, yet likewise omit explicit knowledge resources.
\citet{kbl} introduce reasoning-focused tasks in the Korean legal domain, but these are largely confined to relatively simple yes/no classifications (e.g., causal reasoning, statement consistency, case relevance), and their Korean Bar Exam MCQA does not include related precedents, making it difficult to disentangle reasoning ability from prior knowledge.
\citet{lexeval} offer a large-scale and comprehensive evaluation of Chinese legal knowledge and reasoning, but the benchmark is predominantly multiple-choice–based and does not explicitly evaluate statute-/precedent-grounded faithfulness at the instance level.
Focusing on retrieval, \citet{reasoningfocused} propose legal RAG benchmarks that highlight the reasoning required to formulate retrieval queries, yet their tasks remain limited to multiple-choice and binary questions and are restricted to U.S. legal materials.
Finally, while \citet{koblex} and \citet{LAR-ECHR} move beyond simple classification, addressing provision-grounded open-ended QA with fidelity-aware evaluation and legal argument continuation in ECHR cases, respectively, their scope remains constrained: the former focuses on statute-centric reasoning without precedent-based analysis, and the latter formulates legal reasoning as multiple-choice next-statement prediction rather than fully grounded end-to-end reasoning.

Going beyond prior benchmarks, our goal is to enable the evaluation of reasoning capabilities independently of parameterized knowledge, thereby providing more meaningful insights for LLM development. 
To this end, we introduce the Korean Canonical Legal Benchmark (KCL), a Korean Bar Exam-based benchmark that provides question-level supporting precedents.
KCL further distinguishes itself in two ways: it enables the automatic evaluation of open-ended generation with rubric-based LLM-as-a-Judge, and it targets Korean, a low-resource language, thereby broadening the scope of legal LLM evaluation beyond English-centric contexts.

\section{KCL Dataset Construction}

KCL consists of two complementary benchmarks derived from the Korean Bar Exam: KCL-MCQA, a multiple-choice benchmark, and KCL-Essay, an open-ended generation benchmark (Figure~\ref{fig:KCL}).

\paragraph{Motivation}
KCL is designed to evaluate models along two distinct dimensions.
The first dimension, \texttt{knowledge coverage}, measures how much legal knowledge a model has internally parameterized, which can be assessed directly from question-answer pairs.
The second, \texttt{evidence-grounded reasoning}, captures a model’s ability to apply reasoning when relevant legal knowledge is supplied externally.

Disentangling these two axes is particularly difficult in law, where specialized, country-specific expertise is required and high-quality data remains scarce.
To address this challenge, KCL pairs each question with \texttt{supporting precedents} extracted directly from expert solutions.
These precedents are not answer labels but question-aligned knowledge that provides a foundation for reasoning.

\paragraph{Design} KCL evaluates Large Language Models (LLMs) in two settings: a \texttt{vanilla} setting, where models answer questions without extra resources to reveal their \texttt{knowledge coverage}, and a \texttt{w/ supporting precedents} setting, where \texttt{supporting precedents} are provided to assess \texttt{evidence-grounded reasoning}. Comparing the two allows developers to diagnose and improve knowledge acquisition and reasoning independently, leading to more precise evaluation and better-targeted model development.

\subsection{KCL-MCQA}

We constructed the multiple-choice set by curating questions from the 2024 and 2025 Korean Bar Exams.
All answer options in KCL-MCQA are taken verbatim from the official Korean Bar Exam.
A prior study~\cite{kbl} reported that, even under a RAG setting, using only publicly available datasets of judicial precedents and statutes not directly connected to the task yielded only negligible performance gains (Table 2 in~\citep{kbl}).
As a result, it was difficult to assess how models would perform if provided with sufficient knowledge prior to reasoning. 

To address this, we first curated questions by filtering expert commentaries that cited at least one precedent.
We then retrieved the referenced precedents for each question using a Korean legal document search engine\footnote{LBOX, \url{https://lbox.kr/v2}}, and defined them as \texttt{supporting precedents}.
The resulting benchmark, KCL-MCQA, consists of 283 five-choice questions, with an average of 3.9 \texttt{supporting precedents} per question, for a total of 1,103 precedents (Table~\ref{table:kclstats}).

\subsection{KCL-Essay}

To examine LLMs on legal reasoning tasks, we further focused on the case-based questions of the Korean Bar Exam that correspond to the Multistate Essay Exam (MEE) of the U.S. Uniform Bar Exam. 
They consist of open-ended generation questions that require candidates to identify legal issues, apply statutes and precedents, and construct well-reasoned conclusions (IRAC).

To build KCL-Essay, similar to KCL-MCQA, we first collected case-based questions and their expert commentaries from the past five years (2021-2025) of the Korean Bar Exam.
We then curated 169 questions for which a model’s answer would need to reference at least one precedent.
And we retrieved \texttt{supporting precedents} again.
On average, we obtained 3.3 precedents per question, resulting in a total of 550 precedents (Table~\ref{table:kclstats}). 

Unlike KCL-MCQA, KCL-Essay is an open-ended generation task, and we therefore adopted an LLM-as-a-Judge approach for automated evaluation.
Following prior studies~\cite{sedareval, biggen}, we constructed instance-level rubrics to provide finer-grained criteria tailored to the context of each question.

To build the rubrics, we first drafted an initial set for a representative problem in each subject (civil, criminal, public) with assistance from a licensed Korean attorney.
Using these initial rubrics as a one-shot seed, we prompted Gemini 2.5 Flash to generate rubric sets for all problems with the question text and expert solution.
Finally, three licensed Korean attorneys reviewed the 744 generated rubrics, correcting 13 of them (2\%). Given this low error rate, we built a total of 2,739 rubrics automatically.

In the original essay-type Korean Bar Exam, each question is assigned points reflecting its difficulty. 
We preserved these weightings and evenly distributed the total score across its rubrics, yielding a total score of 2,905 points for the entire benchmark.
We report model scores as percentages of the total in the results below.

\begin{table}[tb]
\centering
\scriptsize
\begin{threeparttable}
\resizebox{0.9\columnwidth}{!}{
\begin{tabular}{l | c c}
\toprule
\textbf{Dataset} & \textbf{KCL-MCQA} & \textbf{KCL-Essay} \\
\midrule
 \# Questions
 & 283 & 169 \\
 Mean tokens per question
 & 552 & 745 \\
 Mean \# rubrics per question
 & - & 16.2 \\
 \midrule
 Mean \# supporting precedents
 & 3.9 & 3.3 \\
 Mean tokens per precedent 
 & 4,631 & 3,648 \\
 Max tokens per precedent
 & 13,833 & 749,580 \\
\bottomrule
\end{tabular}
}
\begin{tablenotes}
\scriptsize
\item[1] \url{https://github.com/openai/tiktoken}
\end{tablenotes}
\caption{Statistics of KCL-MCQA and KCL-Essay.\textsuperscript{1}}
\label{table:kclstats}
\end{threeparttable}
\vspace{-10pt}
\end{table}

\section{Experimental Setup}

\paragraph{Inference} We systematically evaluated 30+ models across the two components, KCL-MCQA and KCL-Essay.
Following \citet{lexam}, we categorized the models into four types: \texttt{Large Reasoning}, \texttt{Small Reasoning}, \texttt{Large General}, and \texttt{Small General}.
For open-API models, since detailed information was not disclosed, we classified them as either \texttt{Small} or \texttt{Large} according to whether they were the smaller or larger variant within the same release family.
For open-weight models, we use 32B parameters for dense models and 120B for MoE models as thresholds to classify them as \texttt{Large} or \texttt{Small}.
Models trained to reason in the verbal space were categorized as \texttt{Reasoning} models.
We used vLLM~\cite{vllm} to evaluate open-weight models, and the usage details for open-API models are provided in Appendix~\ref{app:detail}.
Beyond well-known models, we additionally evaluate LLMs developed in Korea, including HCX SEED~\cite{hcx}, Kanana~\cite{kanana}, A.X~\cite{ax40}, and Exaone~\cite{exaone40}, along with multilingual-focused models such as Aya Expanse~\cite{aya}

\paragraph{Evaluation} We evaluated KCL-Essay using Gemini 2.5 Flash as the judge model. In preliminary comparisons (see Figure~\ref{figure:correlation}), larger judge models showed slightly higher correlations, but Gemini 2.5 Flash was selected for its favorable cost and latency profile while retaining comparable agreement with human experts. The use of instance-level rubrics yielded a strong correlation between human experts and LLM: $\sim$0.9 for questions with total scores below 20 points (80\% of questions), $\sim$0.7 for those above 20 points, and $\sim$0.8 overall (Pearson; Figure~\ref{figure:correlation} in Appendix). Assuming that the original total scores from the bar exam reflect the true difficulty of the questions, these results suggest that while LLM-as-a-Judge with instance-level rubrics is generally effective, caution is needed when evaluating complex examples. Further details are explained in the Appendix.

\begin{table}[tb!]
\centering
\resizebox{0.9\linewidth}{!}{
\begin{tabular}{c | @{}c | c@{}}
\toprule
\multicolumn{3}{c}{\textbf{KCL-MCQA}} \\
\midrule
\makecell[c]{\textbf{Type}}
 & \makecell[c]{\textbf{Model Name}}
 & \makecell[c]{
        \textbf{score (\%, $\uparrow$)} \\
        \textbf{vanilla $\rightarrow$ w/ supporting precedents}
    }
 \\
\midrule
\multirow[c]{10}{*}{\rotatebox{90}{\textbf{Large Reasoning}}}
 & Gemini 2.5 Pro
 & $\mathbf{66.1_{\pm1.0}\rightarrow 89.3_{\pm0.9}}$ \\
 & Claude Opus 4.1\textsuperscript{*}
 & $57.2_{\pm1.0}\rightarrow 83.3_{\pm0.7}$ \\
 & Claude Sonnet 4.5
 & $54.9_{\pm0.7}\rightarrow 85.2_{\pm1.3}$ \\
 & GPT-5
 & $53.9_{\pm1.1}\rightarrow 84.2_{\pm0.2}$ \\
 & Claude 3.7 Sonnet
 & $50.4_{\pm1.0}\rightarrow 82.0_{\pm0.3}$ \\
 & o3
 & $50.2_{\pm1.2}\rightarrow 83.4_{\pm1.8}$ \\
 & Claude Sonnet 4\textsuperscript{*}
 & $45.7_{\pm2.1}\rightarrow 78.6_{\pm1.6}$ \\
 & Qwen3 235B Thinking
 & $41.5_{\pm1.3}\rightarrow 76.9_{\pm1.4}$ \\
 & DeepSeek-R1\textsuperscript{*}
 & $40.9_{\pm1.9}\rightarrow 72.4_{\pm1.3}$ \\
 \cmidrule(r){2-3}
 & \textbf{Average}
 & $51.2\rightarrow81.7$ \\
\midrule
\multirow[c]{8}{*}{\rotatebox{90}{\textbf{Small Reasoning}}}
 & Gemini 2.5 Flash
 & $\mathbf{51.8_{\pm0.6}\rightarrow 83.7_{\pm0.0}}$ \\
 & GPT-5 Mini
 & $40.9_{\pm1.3}\rightarrow 76.8_{\pm1.4}$ \\
 & o4-mini
 & $37.7_{\pm1.2}\rightarrow 74.6_{\pm0.8}$ \\
 & Exaone 4.0 32B
 & $36.4_{\pm1.6}\rightarrow 72.3_{\pm2.2}$ \\
 & Qwen3 32B
 & $35.0_{\pm3.2}\rightarrow 65.4_{\pm1.3}$ \\ 
 & HCX SEED Think 14B\textsuperscript{*}
 & $31.3_{\pm0.8}\rightarrow 57.6_{\pm0.3}$ \\
 & gpt-oss 120B
 & $29.2_{\pm1.6}\rightarrow 64.1_{\pm1.2}$ \\
 \cmidrule(r){2-3}
 & \textbf{Average}
 & $37.5\rightarrow70.6$ \\
\midrule
\multirow[c]{9}{*}{\rotatebox{90}{\textbf{Large General}}}
 & A.X 4.0
 & $\mathbf{49.1_{\pm0.5}}\rightarrow 71.3_{\pm0.4}$ \\
 & GPT-4.1
 & $47.9_{\pm1.0}\rightarrow 72.0_{\pm0.4}$ \\
 & Claude 3.5 Sonnet\textsuperscript{*}
 & $45.1_{\pm0.3}\rightarrow 70.8_{\pm0.9}$ \\
 & Llama 4 Maverick
 & $43.3_{\pm0.4}\rightarrow 72.3_{\pm1.0}$ \\
 & GPT-4o
 & $43.3_{\pm2.0}\rightarrow 63.8_{\pm1.0}$ \\
 & Qwen3 235B Instruct
 & $40.9_{\pm1.3}\rightarrow \mathbf{74.2_{\pm0.8}}$ \\
 & Kimi K2 Instruct
 & $39.5_{\pm1.2}\rightarrow 71.6_{\pm1.4}$ \\
 & DeepSeek-V3
 & $37.3_{\pm0.7}\rightarrow 69.3_{\pm0.3}$ \\
 \cmidrule(r){2-3}
 & \textbf{Average}
 & $43.3\rightarrow70.7$ \\
\midrule
\multirow[c]{8}{*}{\rotatebox{90}{\textbf{Small General}}}
 & Gemma 3 27B
 & $\mathbf{35.1_{\pm0.3}}\rightarrow 53.6_{\pm0.4}$ \\
 & GPT-4o-mini
 & $29.9_{\pm1.5}\rightarrow 46.4_{\pm0.9}$ \\
 & GPT-4.1-mini
 & $29.4_{\pm0.7}\rightarrow \mathbf{60.5_{\pm1.2}}$ \\
 & Kanana 1.5 8B
 & $29.4_{\pm0.8}\rightarrow 47.5_{\pm1.4}$ \\
 & Aya Expanse 32B
 & $28.4_{\pm0.2}\rightarrow 42.3_{\pm0.4}$ \\
 & Claude 3.5 Haiku\textsuperscript{*}
 & $28.0_{\pm1.3}\rightarrow 54.5_{\pm1.4}$ \\
 & Phi-4-mini
 & $22.0_{\pm0.8}\rightarrow 30.6_{\pm0.4}$ \\
 \cmidrule(r){2-3}
 & \textbf{Average}
 & $28.9\rightarrow47.9$ \\
\midrule
\multicolumn{2}{c|}{Random} & 20 \\
\bottomrule
\end{tabular}
}
\caption{Evaluation results on KCL-MCQA, averaged over three independent runs. Precedents exceeding the maximum context length were truncated (indicated by *). "Average" denotes the average performance within each group.}
\label{table:choicefull}
\vspace{-10pt}
\end{table}

\begin{table}[tb!]
\centering
\resizebox{0.9\linewidth}{!}{
\begin{tabular}{c | @{}c | c@{}}
\toprule
\multicolumn{3}{c}{\textbf{KCL-Essay}} \\
\midrule
\makecell[c]{\textbf{Type}}
 & \makecell[c]{\textbf{Model Name}}
 & \makecell[c]{
        \textbf{score (\%, $\uparrow$)} \\
        \textbf{vanilla $\rightarrow$ w/ supporting precedents}
    }
 \\
\midrule
\multirow[c]{10}{*}{\rotatebox{90}{\textbf{Large Reasoning}}}
 & Gemini 2.5 Pro
 & $\mathbf{60.5_{\pm0.5}\rightarrow 74.9_{\pm0.5}}$ \\
 & GPT-5
 & $57.5_{\pm1.3}\rightarrow 72.8_{\pm0.3}$ \\
 & o3
 & $51.2_{\pm0.6}\rightarrow 69.4_{\pm0.8}$ \\
 & Claude Sonnet 4.5
 & $47.6_{\pm1.5}\rightarrow 68.5_{\pm0.4}$ \\
 & Claude Opus 4.1\textsuperscript{*}
 & $45.5_{\pm0.2}\rightarrow 60.3_{\pm0.5}$ \\
 & Claude 3.7 Sonnet
 & $43.6_{\pm0.7}\rightarrow 63.6_{\pm0.8}$ \\
 & Qwen3 235B Thinking
 & $39.2_{\pm1.1}\rightarrow 63.9_{\pm0.1}$ \\
 & Claude Sonnet 4\textsuperscript{*}
 & $35.5_{\pm1.2}\rightarrow 54.9_{\pm0.2}$ \\
 & DeepSeek-R1\textsuperscript{*}
 & $35.0_{\pm0.5}\rightarrow 53.1_{\pm0.2}$ \\
 \cmidrule(r){2-3}
 & \textbf{Average}
 & $46.2\rightarrow64.6$ \\
\midrule
\multirow[c]{8}{*}{\rotatebox{90}{\textbf{Small Reasoning}}}
 & Gemini 2.5 Flash
 & $\mathbf{53.7_{\pm1.1}\rightarrow 71.1_{\pm0.1}}$ \\
 & GPT-5 Mini
 & $40.3_{\pm0.5}\rightarrow 65.5_{\pm0.5}$ \\
 & Exaone 4.0 32B
 & $32.6_{\pm0.4}\rightarrow 47.5_{\pm1.1}$ \\
 & gpt-oss 120B
 & $31.6_{\pm0.5}\rightarrow 56.0_{\pm0.3}$ \\
 & o4-mini
 & $31.3_{\pm0.9}\rightarrow 54.5_{\pm0.3}$ \\
 & Qwen3 32B
 & $29.2_{\pm0.3}\rightarrow 55.4_{\pm1.2}$ \\
 & HCX SEED Think 14B\textsuperscript{*}
 & $26.2_{\pm0.2}\rightarrow 42.8_{\pm0.7}$ \\
 \cmidrule(r){2-3}
 & \textbf{Average}
 & $35.0\rightarrow56.1$ \\
\midrule
\multirow[c]{9}{*}{\rotatebox{90}{\textbf{Large General}}}
 & GPT-4.1
 & $\mathbf{42.4_{\pm0.8}\rightarrow 65.1_{\pm0.6}}$ \\
 & Qwen3 235B Instruct
 & $38.9_{\pm0.9}\rightarrow 63.0_{\pm0.5}$ \\
 & A.X 4.0
 & $34.7_{\pm0.6}\rightarrow 52.4_{\pm0.9}$ \\
 & DeepSeek-V3
 & $34.3_{\pm0.6}\rightarrow 57.6_{\pm1.5}$ \\
 & Claude 3.5 Sonnet\textsuperscript{*}
 & $33.7_{\pm0.3}\rightarrow 50.3_{\pm0.7}$ \\
 & Kimi K2 Instruct
 & $33.3_{\pm0.3}\rightarrow 55.7_{\pm0.6}$ \\
 & Llama 4 Maverick
 & $25.6_{\pm0.3}\rightarrow 44.3_{\pm0.8}$ \\
 & GPT-4o
 & $24.3_{\pm0.4}\rightarrow 42.0_{\pm0.1}$ \\
 \cmidrule(r){2-3}
 & \textbf{Average}
 & $33.4\rightarrow53.8$ \\
\midrule
\multirow[c]{8}{*}{\rotatebox{90}{\textbf{Small General}}}
 & GPT-4.1-mini
 & $\mathbf{30.9_{\pm0.7}\rightarrow 56.1_{\pm1.0}}$ \\
 & Gemma 3 27B
 & $23.6_{\pm0.4}\rightarrow 44.0_{\pm0.5}$ \\
 & Kanana 1.5 8B
 & $21.8_{\pm0.8}\rightarrow 49.1_{\pm0.4}$ \\
 & Claude 3.5 Haiku\textsuperscript{*}
 & $21.2_{\pm0.6}\rightarrow 39.9_{\pm0.1}$ \\
 & Aya Expanse 32B
 & $20.2_{\pm0.3}\rightarrow 37.2_{\pm0.1}$ \\
 & GPT-4o-mini
 & $16.1_{\pm0.3}\rightarrow 32.6_{\pm0.4}$ \\
 & Phi-4-mini
 & $6.0_{\pm0.3}\rightarrow 18.9_{\pm0.1}$ \\
 \cmidrule(r){2-3}
 & \textbf{Average}
 & $20.0\rightarrow39.7$ \\
\bottomrule
\end{tabular}
}
\caption{Evaluation results on KCL-Essay. The experimental settings are identical to those in Table~\ref{table:choicefull}.}
\label{table:essay}
\vspace{-10pt}
\end{table}

\section{Analysis}

\subsection{The Impact of Legal Knowledge on Problem Solving Performance}

We evaluated various LLMs on the \texttt{vanilla} setting first, where supporting precedents are not provided, to examine differences in parameterized knowledge across models.

As shown in Table~\ref{table:choicefull}, performance is strongly associated with model size. 
The average performance followed the order of \texttt{Large Reasoning}, \texttt{Large General}, \texttt{Small Reasoning}, and \texttt{Small General}.
This accords with findings from prior work that larger models tend to memorize more during training~\cite{memorizescale}.
Among \texttt{Large Reasoning} models, Gemini 2.5 Pro performs highest overall, followed by Claude and GPT-5 effectively showing their parameterized legal knowledge in a low-resource language such as Korean.

Next, in the \texttt{w/ supporting precedents} setting, where corresponding precedents were provided as in-context input, scores increased substantially compared to the \texttt{vanilla} setting across all models.
Interestingly, the average performance gap between the \texttt{Large General} and \texttt{Small Reasoning} types narrows to just 0.1\%p. 
This suggests that while \texttt{Small Reasoning} models have lower \texttt{knowledge coverage} than \texttt{Large General} models, they may nevertheless demonstrate superior \texttt{evidence-grounded reasoning}.
Also note that many models surpass the human passing threshold (68.7\%~\cite{superlawyer}) underscoring that high-quality legal data remains scarce.

\subsection{On the Importance of Reasoning for KCL-Essay}
\label{section:reasoning}
Unlike KCL-MCQA, in KCL-Essay, the model exhibits average performance in the order of \texttt{Large Reasoning}, \texttt{Small Reasoning}, \texttt{Large General}, and \texttt{Small General} in both \texttt{vanilla} and \texttt{w/ supporting precedents} settings.
Furthermore, GPT-5 and o3 outperform the Claude model family across all settings.
These results suggest that KCL-Essay requires deeper reasoning ability. 

Nevertheless, even in the \texttt{w/ supporting precedents} setting, the absolute scores remained relatively low, unlike in KCL-MCQA. 
Even the best-performing model, Gemini 2.5 Pro, still exhibited a remaining gap of about 25\%p, underscoring the need for further advances in legal reasoning.

\subsection{On the Performance of Open-Weight Models in KCL}
\label{app:ow}

We analyze the performance of open-weight models. 
As shown in Table~\ref{table:choicefull}, for KCL-MCQA, it is noteworthy that in the \texttt{Large Reasoning} and \texttt{Small Reasoning} types, the best-performing models across all settings were open-API models.
By contrast, in the \texttt{Large General} and \texttt{Small General} types, the best models in the \texttt{vanilla} setting were the open-weight models A.X 4.0~\cite{ax40} and Gemma 3 27B~\cite{gemma3}, respectively, although they fell behind open-API models in the \texttt{w/ supporting precedents} setting. 
This suggests that open-weight models may achieve competitive performance in terms of \texttt{knowledge coverage} within certain vertical domains, while still showing limitations in deeper reasoning.

For KCL-Essay, as shown in Table~\ref{table:essay}, the top-performing model in every type was consistently an open-API model. 
Although open-weight LRMs have demonstrated strength in STEM and code-related benchmarks~\cite{r1}, they continue to face clear challenges in specialized domains requiring expert-level reasoning, where API-based models hold a substantial advantage.
These results highlight the value of KCL as a reliable reference for advancing the open-weight model ecosystem.

\subsection{Effects of Language Specialization}

Given that our benchmark data is based on Korean, a low-resource language, we also conducted a closer analysis of models specialized in Korean.
In KCL-MCQA, as shown in Table~\ref{table:choicefull}, HCX SEED Think 14B achieved higher performance in the \texttt{vanilla} setting than the recent open-weight LLM gpt-oss 120B. 
Similarly, as noted in Section~\ref{app:ow}, A.X 4.0 surpassed all other open-API and open-weight models in the \texttt{Large General} type. 
However, both models exhibited limited improvement in the \texttt{w/ supporting precedents} setting and were eventually outperformed by other models.

In KCL-Essay, interestingly, Exaone 4.0 32B and A.X 4.0, both trained specifically for Korean, even outperformed gpt-oss 120B and Kimi K2 Instruct in the \texttt{vanilla} setting, but underperformed relative to them in the \texttt{w/ supporting precedents} setting.
Although these models benefit from Korean-specific knowledge in terms of \texttt{knowledge coverage}, their reasoning ability still lags behind, underscoring the need for further advances in reasoning-focused training.

In contrast to KCL-MCQA, in KCL-Essay the HCX SEED Think 14B~\cite{hcx}, despite being specifically trained for Korean, recorded the lowest score among models in the \texttt{Small Reasoning} category across all settings.
This supports our earlier finding in Section~\ref{section:reasoning} that KCL-Essay places a stronger emphasis on expert-level reasoning rather than knowledge coverage alone.

Taken together, this analysis underscores the importance of KCL as a benchmark in the broader context of developing national AI strategies.

\section{Conclusion}

We propose KCL for knowledge-independent legal reasoning: KCL-MCQA, with 283 questions and 1,103 aligned precedents, and KCL-Essay, with 169 problems, 550 precedents, and 2,739 rubrics.
With \texttt{supporting precedents}, many frontier LRMs surpass 80\% on KCL-MCQA; without them, errors largely reflect missing knowledge.
KCL-Essay remains difficult: the best score is 74.9\% even with precedents.
KCL’s \texttt{supporting precedents} and instance-level rubrics enable controlled, reproducible evaluation.
Overall, KCL is designed to foster models that \emph{reason}, rather than merely \emph{recall} knowledge, by disentangling \texttt{evidence-grounded reasoning} from parameterized knowledge when evaluating model capabilities.

\section*{Limitation}

KCL focuses exclusively on Korean legal tasks, limiting direct applicability to other legal systems and languages. 
For KCL-Essay, we rely on LLM-as-a-Judge (Gemini 2.5 Flash) for automated scoring, which, despite validated correlation with human expert evaluations, may not capture all nuances of legal reasoning quality.
The \texttt{w/ supporting precedents} setting assumes perfect knowledge retrieval and does not fully isolate the retrieval-reasoning hybrid challenge of identifying which parts of provided precedents are relevant. 
Future work on KCL-Essay could benefit from aligning the evaluation rubric with the IRAC framework, which would allow disentangling and assessing reasoning at each stage more explicitly.

Another limitation is that our benchmark evaluates model performance on questions that are already annotated with correct precedents in expert solutions.
In real-world deployments, legal AI systems are typically implemented as retrieval-augmented generation (RAG) pipelines.
Because our study does not model an end-to-end pipeline that performs retrieval and reasoning jointly, we do not account for noise, omissions, or misranking introduced during retrieval.
Nevertheless, the benchmark explicitly provides question-aligned precedents, which enables an additional and complementary use case: the annotated precedents serve as gold references, allowing retrieval components to be evaluated.

\section*{Acknowledgments}
We thank Sooyung Park, a licensed lawyer in South Korea, for valuable advice on handling legal data.
Hongseok Oh and Wonseok Hwang were supported by the National Research Foundation of Korea (NRF) grant funded by the Korea government(MSIT) (RS-2025-23524855).

\bibliography{main}

\newpage

\appendix
\section{Evaluation}
\subsection{Reliability of the LLM-as-a-Judge in KCL-Essay}

As a preliminary consistency check, we first scored the expert solution against the finalized rubrics.
When evaluating the expert solution as the ground truth against the finalized rubric set, we obtained a mean score of $99.2_{\pm0.1}\%$ over three independent runs, confirming near-perfect alignment between the rubric and the ground truth answers.

\begin{figure}[h!]
    \centering
    \includegraphics[width=\linewidth]{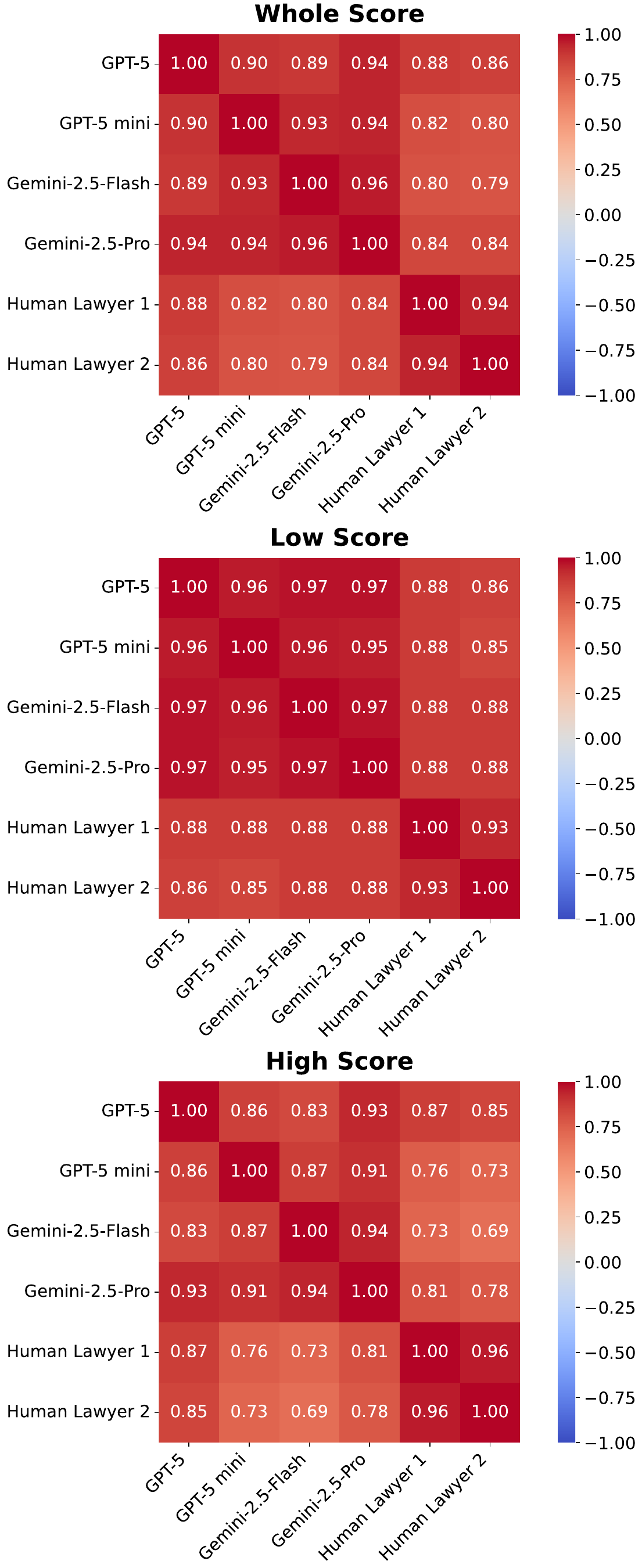}
    \caption{Pearson correlation heatmaps between LLMs and human expert evaluators. Whole Score represents the correlation over all question-prediction pairs derived from 641 triplets, while Low Score corresponds to the correlation for easy questions (bottom 80\%), and High Score corresponds to the correlation for difficult questions (top 20\%).}
    \label{figure:correlation}
\end{figure}

To further assess reliability, we compared the evaluation results of Gemini with those of human experts.
We randomly sampled 641 question-prediction-rubric triplets, corresponding to 41 question-model pairs from the full results in Table~\ref{table:essay}.
Two independent groups of three licensed Korean attorneys each scored all samples independently.
As shown in Figure~\ref{figure:correlation}, we obtained an overall Pearson correlation of about 0.8 (top of the Figure~\ref{figure:correlation}) between Gemini 2.5 Flash and human evaluators.

We further clustered the questions into two groups according to their original bar exam scores, which reflect difficulty levels.
First, to determine the cutoff for the top 20\%, we examined the score distribution of whole KCL-Essay questions, and found that a score of 20 served as the threshold (Fig.~\ref{figure:distribution} in Appendix).
Using this threshold, we classified the 41 question-model pairs with human evaluations into two categories: those with scores up to 20 were labeled as Easy, and those above 20 as Hard.
A clear difference emerged: for easier questions (bottom 80\%), both the inter-human correlations 0.93 and the human-LLM correlations (0.88) were high.
For more difficult questions, however, the correlations dropped substantially: the Gemini-human Pearson correlation was around 0.7, while the inter-human correlation also decreased to about 0.96.
These findings suggest that for harder questions, the use of LLM-as-a-Judge requires more careful consideration.

\section{LICENSE}

\subsection{Model}

GPT-5, GPT-5 Mini~\cite{gpt5} and GPT 4.1~\cite{gpt41}, o3~\cite{o3}, GPT-4o and GPT-4o-mini~\cite{gpt4o} were used in compliance with the OpenAI Terms of Use~\cite{oaiterm}.
Gemini 2.5 Pro~\cite{gemini25pro}, Gemini 2.5 Flash~\cite{gemini25flash} were used in compliance with the Google APIs Terms of Service~\cite{geminiterm}.
Claude Opus 4.1~\cite{claudeopus41}, Claude Sonnet 4.5~\cite{claudesonnet45}, Claude Sonnet 4~\cite{claudesonnet4}, Claude 3.7 Sonnet~\cite{claude37} were used in compliance with the Anthropic on Bedrock - Commercial Terms of Service~\cite{anthropicbedrock}. 
DeepSeek-V3~\cite{deepseekv3} is available under DEEPSEEK LICENSE~\cite{deepseekv3license}.
DeepSeek-R1~\cite{r1} and Phi-4-mini~\cite{phi4} are available under MIT License.
Qwen3 235B Thinking, Qwen3 235B, Qwen3 32B Instruct~\cite{qwen3}, gpt-oss 120B~\cite{gptoss}, and Kanana 1.5 8B~\cite{kanana} are available under Apache License Version 2.0.
Exaone 4.0 32B~\cite{exaone40} is available under EXAONE AI Model License Agreement 1.2 - NC.
A.X 4.0~\cite{ax40} is under Qwen LICENSE.
Kimi K2 Instruct~\cite{kimik2} is available under Modified MIT LICENSE~\cite{kimik2license}.
Llama 4 Maverick~\cite{llama4} is available under LLAMA 4 COMMUNITY LICENSE~\cite{llama4license}.
Gemma 3 27B~\cite{gemma3} is available under the Gemma Terms of Use~\cite{gemmaterms}
HCX SEED Think 14B~\cite{hcx} is available under the HyperCLOVA X SEED 14B Think Model License Agreement~\cite{hcxlicense}.
Aya Expanse 32B~\cite{aya} is available under the Creative Commons Attribution 4.0 License (CC BY 4.0).

We used the models in accordance with their respective licenses.

\subsection{Data}

The Korean Bar Exam is produced by the Ministry of Justice of Korea and distributed under Type 1 of the Korean Open Government License (KOGL), which permits both commercial and non-commercial use provided that the source is credited. Court judgments in Korea are classified as works not subject to copyright protection under Article 7(3) of the Korean Copyright Act and are therefore freely available for use.

\subsection{KCL}

We release our KCL data and evaluation code under the CC BY-NC 4.0 license.

\section{Details of Model Usage}
\label{app:detail}

In this section, we clarify the versions and configurations of the models referenced in the main text.
When using Gemini 2.5 Pro and Gemini 2.5 Flash, we utilized Vertex AI’s gemini-2.5-pro and gemini-2.5-flash, respectively, with dynamic thinking enabled.
The same configuration was used when employing Gemini 2.5 Flash as an LLM-as-a-Judge.
The GPT-5 mentioned in the main text refers to gpt-5-2025-08-07, GPT-5 Mini refers to gpt-5-mini-2025-08-07, o4-mini refers to o4-mini-2025-04-16, and o3 refers to o3-2025-04-16, GPT-4o refers to gpt-4o-2024-08-06, all accessed through the OpenAI API, with reasoning effort set to medium.
GPT-4.1 and GPT-4.1-mini were also accessed through the OpenAI API, specifically gpt-4.1-2025-04-14 and gpt-4.1-mini-2025-04-14, respectively.
Claude Opus 4.1, Claude Sonnet 4.5, Claude Sonnet 4, and Claude 3.7 Sonnet were accessed via Amazon Bedrock using us.anthropic.claude-opus-4-1-20250805-v1:0, us.anthropic.claude-sonnet-4-5-20250929-v1:0, us.anthropic.claude-sonnet-4-20250514-v1:0, and us.anthropic.claude-3-7-sonnet-20250219-v1:0, us.anthropic.claude-3-5-sonnet-20240620-v1:0, us.anthropic.claude-3-5-haiku-20241022-v1:0 respectively, with the thinking budget set to 8k when using reasoning mode.
DeepSeek-R1 and Llama 4 Maverick were also accessed via Amazon Bedrock using us.deepseek.r1-v1:0 and us.meta.llama4-maverick-17b-instruct-v1:0, respectively.
All other open-weight models were run using vLLM, and GPU usage amounted to approximately 100 H100 hours in total on the cloud.

\section{Labeling Detail}

\subsection{Rubric Extraction Prompt}

Table~\ref{table:rubricextract} shows the prompt used for rubric extraction in English (translated).  
We utilized Gemini 2.5 Flash to extract the rubrics, and expert review was completed as described in the main text.

\begin{table}[tbh]
\centering
\begin{tabular}{p{0.95\linewidth}}
\toprule
\textbf{Rubric Extract Prompt} \\
\midrule
\ttfamily
Please create [rubrics] for grading the [submitted answer sheet] based on the given [model answer] from the Korean Bar Exam commentary.  

The generated [rubrics] must allow checking the following points in the [submitted answer sheet]:  
1. Whether all key issues are addressed (Issue).  
2. Whether the relevant legal principles, major statutes, or conclusions from key precedents are cited (Rule).  
3. Whether the legal principles, statutes, and major precedents are properly applied to the given case (Application) and explained logically.  
4. Whether the same conclusion (Conclusion) as in the [model answer] is reached.  

Please avoid creating grading criteria that only check for formal compliance without substantive content.  
- **Bad rubric example**: ...(omitted)...  
- **Good rubric example**: ...(omitted)...  

Also, when citing key precedents, do not include the exact case number in the rubric. However, it is important to require mentioning the relevant content.  
- **Bad rubric example**: ...(omitted)...  
- **Good rubric example**: ...(omitted)...  

Please refer to the following transformation example when writing:  

[Example Question]: ...(omitted)...  
[Example Rubric]: ...(omitted)...  

[Question]: ...(omitted)...  
[GT Answer]: ...(omitted)...     \\
\bottomrule
\end{tabular}
\caption{Rubric Extraction Prompt}
\label{table:rubricextract}
\end{table}

\subsection{Annotator Instructions}
We collaborated with licensed attorneys using Google Sheets, and provided separate instructions for rubric review and model output evaluation, as depicted in Table~\ref{table:rubricinstruction} and Table~\ref{table:evalinstruction} in English (translated).

\begin{table}[tbh]
\centering
\begin{tabular}{p{0.95\linewidth}}
\toprule
\textbf{Rubric Evaluation Instruction} \\
\midrule
\ttfamily
Based on the review sheet, please evaluate the appropriateness of the evaluation rubric generated by AI, propose additional rubric if necessary, and provide an overall opinion on their quality.
1. Assessment of Individual Evaluation Criteria (O/X)
- The sheet [XX Evaluation Criteria Review] contains, for each question from the Korean Bar Exam case questions (Public Law, Criminal Law, Civil Law), the following: “Question Number,” “Question,” “Model Answer (citing Union and Rainbow textbooks),” “Referenced Cases (cases cited in the model answer),” and the evaluation criteria generated by AI.  
- Please determine whether each evaluation rubric is concrete and clear. If so, mark with an **O**; if not, provide a reason. If the criterion itself is inappropriate, also provide the reason.  

2. Proposal of Additional Evaluation Rubric 
- While reviewing the evaluation criteria in the “Evaluation Criteria Review” sheet, please check if any important issues are missing for each question. If something is missing, please add it in the “Additional Evaluation Criteria” column.

3. Overall Opinion on the Quality of Evaluation Criteria
- In the “Overall Opinion” sheet, please provide a general qualitative assessment of how good the evaluation criteria generated by judge AI are. A qualitative review is required rather than numerical scoring.  \\
\bottomrule
\end{tabular}
\caption{Rubric Evaluation Instruction}
\label{table:rubricinstruction}
\end{table}

\begin{table}[tbh]
\centering
\begin{tabular}{p{0.95\linewidth}}
\toprule
\textbf{Model Output Evaluation Instruction} \\
\midrule
\ttfamily
Each sheet contains the following columns: question (the exam problem), pred (the AI’s answer), and rubric (the evaluation criteria).
The problems are drawn from actual Korean bar exam case questions. The AI answers are provided by various models (both strong and weak). The Rubric is designed to resemble an official scoring guideline.  

The evaluator’s task is to read the problem and the AI’s answer, then decide whether the AI’s answer meets the Rubric. The evaluation should be binary (0 = does not satisfy, 1 = satisfies).

The Google Sheet contains two identical sheets. Each sheet must be completed by a different attorney. In other words, an attorney who evaluates Sheet 1 should not participate in evaluating Sheet 2. \\
\bottomrule
\end{tabular}
\caption{Model Output Evaluation Instruction}
\label{table:evalinstruction}
\end{table}

\subsection{Annotator Cost}

In this section, we discuss the cost associated with expert labeling used in the construction of our benchmark. We engaged licensed Korean lawyers under consulting contracts to conduct rubric reviews and output evaluations. A total of six attorneys participated, and they were compensated at a rate of more than \$200 per hour.

\section{Additional Stats on KCL}

\begin{figure}[tbh]
\centering
    \includegraphics[width=\linewidth]{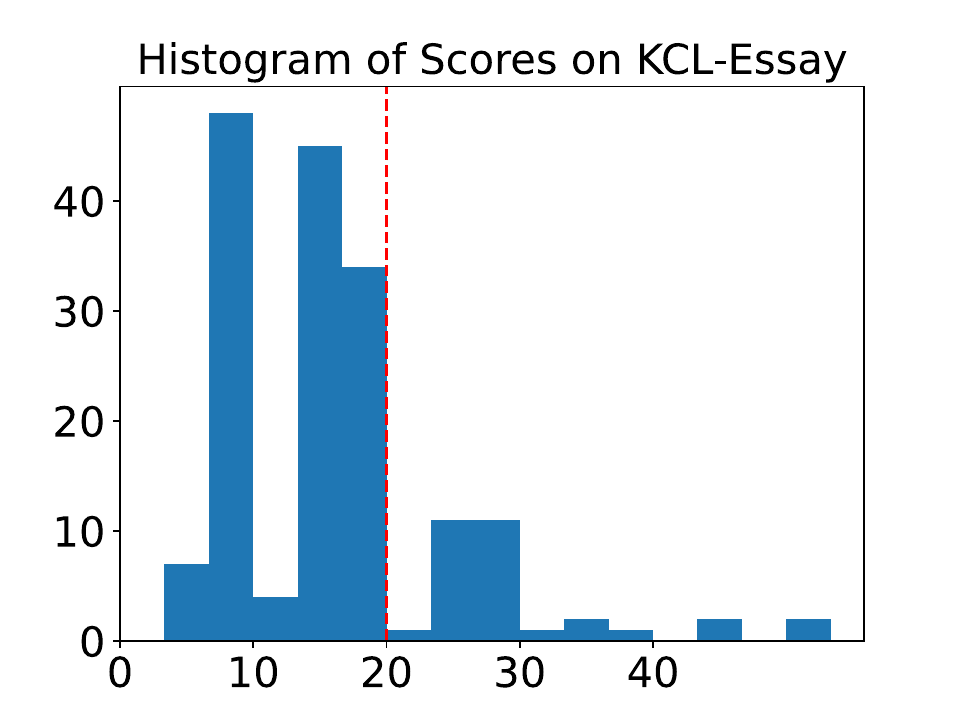}
    \caption{Histogram of score distribution across all 169 questions in KCL-Essay. The cutoff for the top 20\% is observed at a score of 20.}
    \label{figure:distribution}
\end{figure}

As shown in Figure~\ref{figure:distribution}, the distribution of question scores in KCL-Essay is highly skewed, with most questions concentrated below a score of 20. 
To distinguish between relatively easy and more challenging questions in our correlation analysis, we set the top 20\% of questions by score (cutoff 20) as the hard subset. 
This stratification provides a principled way to examine whether the reliability of LLM-as-a-Judge differs depending on question difficulty, rather than relying on an arbitrary threshold.

\end{document}